\documentclass[letterpaper]{article} 
\usepackage[]{aaai2027}  
\usepackage[hyphens]{url}  
\usepackage{graphicx} 
\usepackage{natbib}  
\usepackage{caption} 
\usepackage{algorithm}
\usepackage{algorithmic}
\usepackage{multirow}
\usepackage{newfloat}
\usepackage{listings}
\DeclareCaptionStyle{ruled}{labelfont=normalfont,labelsep=colon,strut=off} 
\floatstyle{ruled}
\newfloat{listing}{tb}{lst}{}
\floatname{listing}{Listing}

\usepackage{booktabs}
\usepackage{amsmath}
\usepackage{amssymb}

\nocopyright

\title{USP-Mamba: Unmixing-Derived Spectral and Structural Prompting for Hyperspectral Image Super-Resolution}
\author {
     Shi Chen,
     Jie Zhang,
     Yicong Zhou\corresponding
}
\affiliations{
    Department of Computer Science, University of Macau, Macau, China\\
    chenshi@um.edu.mo
}

\begin{document}

\maketitle

\begin{abstract}
Hyperspectral image super-resolution aims to reconstruct high-resolution imagery while faithfully preserving dense spectral information. Recently, Mamba-based models have shown promising potential for this task by capturing long-range dependencies with linear computational complexity. Nevertheless, their causal sequence modeling requires two-dimensional hyperspectral features to be unfolded along predefined scanning orders, which disrupts spatial adjacency and restricts the effective propagation of contextual information. Moreover, the state-space parameterization of existing Mamba models is predominantly derived from generic learned representations, without explicit alignment with the intrinsic characteristics of the hyperspectral image. To address this issue, we propose an Unmixing-derived Spectral and Structural Prompting Mamba framework, termed USP-Mamba, which adapts Mamba state evolution through composition-aware spectral priors and image-dependent structural prompts. Specifically, an unmixing-informed spectral prompt captures the global material composition of the input image and provides persistent conditioning throughout reconstruction. Injected into the Mamba sequence and progressively adapted across layers, it steers state evolution toward composition-consistent reconstruction. We further introduce feature-level structural prompts comprising spatial and frequency components to provide image-dependent local guidance. The spatial prompt promotes structure-sensitive state encoding for local detail preservation, while the frequency prompt enables region-adaptive transitions between homogeneous regions and high-frequency details. Finally, complementary Hilbert and Semantic-Guided Neighboring scans preserve spatial continuity and strengthen non-local semantic dependency modeling, respectively. Extensive experiments on different datasets demonstrate that the proposed method consistently outperforms representative approaches. The source code will be available soon.
\end{abstract}


\section{Introduction}

Hyperspectral images (HSIs) record scene information across numerous contiguous spectral bands, providing both spatial details and rich material-specific spectral signatures~\cite{PR2024, HW2024}. However, the physical constraints of incident energy within hyperspectral imaging systems result in an inherent trade-off between spatial and spectral resolution~\cite{LQ2025,XF2026}. Single hyperspectral image super-resolution (SHSR) aims to reconstruct a high-resolution HSI from a single low-resolution observation while preserving its rich spectral information~\cite{CST}.

Several previous works~\cite{CL2022,XZ2025, WW2025} has shown that long-range spatial–spectral dependencies are critical to SHSR. Transformer-based methods effectively capture long-range spatial–spectral dependencies through self-attention, but their computational and memory costs grow quadratically with the number of tokens, limiting their efficiency for high-dimensional hyperspectral images~\cite{MJ2025}. Selective state space models~\cite{ssm}, represented by Mamba~\cite{mamba}, provide an efficient alternative by modeling long-range dependencies with linear complexity and have recently shown promising performance in image restoration and hyperspectral processing.

Despite these advances, current visual state-space models still face several fundamental limitations. First, Mamba relies on causal sequence modeling, where each token representation is accumulated from preceding pixels along a predefined scanning order~\cite{fmsr}. Flattening two-dimensional HSI features into a one-dimensional sequence disrupts spatial adjacency and weakens local information propagation~\cite{JW2026}. Consequently, fine spatial details such as edges and structural details are easily overlooked during reconstruction. Furthermore, the reconstruction performance can be affected by the selected scanning strategy because different sequence orders establish different causal dependencies~\cite{ZL2026}. Although multi-directional scanning strategies alleviate this issue, they still lack an explicit mechanism to jointly consider local structural characteristics and non-local semantic relationships.

Beyond the limitations of sequence unfolding, existing Mamba-based HSISR methods~\cite{mambahsisr} mainly learn state dynamics from generic image features. Although selective state space models predict input-dependent parameters, their state dynamics are still largely derived from generic learned representations without explicit hyperspectral image-specific guidance. MambaIRv2~\cite{mambairv2} partially addresses this issue by introducing a learnable prompt pool to extend state readout beyond the causal sequence; however, these learnable prompts are optimized from training data, rather than explicitly derived from the intrinsic properties of the input. For HSISR, such generic parameterization may indiscriminately propagate redundant spectral responses, inadequately preserve material-dependent correlations~\cite{ZZ2026}, and apply similar transition dynamics to smooth regions and high-frequency details.

To address these limitations, we propose an Unmixing-derived Spectral and Structural Prompting Mamba framework, termed USP-Mamba, for hyperspectral image super-resolution. USP-Mamba introduces hyperspectral-specific guidance into Mamba through global spectral conditioning and local structural modulation. Specifically, an unmixing-informed spectral prompt summarizes the material composition of the input HSI and serves as a persistent condition throughout the network. By injecting it into the scanning sequence and progressively adapting it across layers, USP-Mamba aligns intermediate representations and state dynamics with global spectral characteristics. To provide local structural guidance, we further introduce structural prompts comprising spatial and frequency components. The spatial prompt facilitates structure-sensitive state encoding to preserve local details, whereas the frequency prompt adjusts transitions of state parameters to the differing characteristics of homogeneous regions and high-frequency details. Together, they improve the selective utilization of latent states for spatial–spectral reconstruction. In addition, complementary Hilbert and Semantic-Guided Neighboring~\cite{mambairv2} scans preserve spatial continuity and strengthen non-local semantic interactions, respectively, while their adaptive fusion integrates local geometry with global context. Extensive experiments on multiple benchmark datasets demonstrate the superiority of USP-Mamba over representative HSISR methods. The main contributions of this paper are summarized as follows:

\begin{itemize}
    \item We propose USP-Mamba, a hyperspectral-specific state space framework that integrates global spectral conditioning and local structural modulation. It aligns Mamba dynamics with the intrinsic spectral and spatial-frequency characteristics of HSIs.
    \item We devise a composition-aware spectral prompting strategy that transforms spectral unmixing priors into persistent guidance for Mamba. By progressively conditioning state evolution with global material composition, it promotes composition-consistent spatial–spectral reconstruction.
    \item We develop spatial and frequency structural prompts to promote structure-sensitive state encoding and region-adaptive state transitions. Complementary Hilbert and Semantic-Guided Neighboring scans further preserve spatial continuity while capturing non-local semantic dependencies.
    \item Extensive experiments on multiple benchmark datasets demonstrate the effectiveness and competitive performance of USP-Mamba.
\end{itemize}

\section{Related Work}
\subsection{Single hyperspectral image super-resolution}
SHSR reconstructs an HR-HSI from a single LR observation without auxiliary images. Although the CNN-based methods effectively exploit local spatial--spectral correlations, their limited receptive fields restrict the modeling of long-range dependencies. Transformer-based methods \cite{sqformer, XF2026b, XZ2026} were therefore introduced to enlarge the receptive field. ESSAformer \cite{essaformer} developed efficient spectral-correlation attention. MSDformer \cite{msdformer} further captured multiscale spatial--spectral dependencies using deformable attention. Nevertheless, self-attention generally incurs high computational costs for high-dimensional HSIs, while restricted or approximated attention may weaken global information interaction. More recently, EigenSR~\cite{su2025eigensr} transfers the pre-trained models to HSIs through eigenimage representations.

\subsection{State Space Models}
State space models (SSMs) \cite{ssm} represent sequential dependencies through latent state transitions, while Mamba \cite{mamba} further makes the state-space parameters input-dependent, enabling selective information propagation with linear complexity. Recent studies have extended Mamba to image restoration and HSISR. MambaIR \cite{mambair} incorporated local enhancement and channel attention to alleviate local pixel forgetting and channel redundancy. MambaIRv2 \cite{mambairv2} introduced attentive state-space modeling and semantic-guided neighboring to reduce the causal restriction of scanned sequences. For hyperspectral reconstruction, MambaHSISR
\cite{mambahsisr} employed separate spatial and spectral Mamba subnetworks, whereas HSRMamba \cite{hsrmamba} used local spatial--spectral partitioning and global spectral reordering to improve contextual dependency modeling. Despite these advances, existing methods mainly focus on architectural or scanning refinements. The state evolution is still driven primarily by generic intermediate features, without explicitly incorporating composition-aware and region-dependent structural characteristics.

\section{Method}
\subsection{Overall Architecture}

\begin{figure*}[t] 
\centering 
\includegraphics[width=\textwidth]{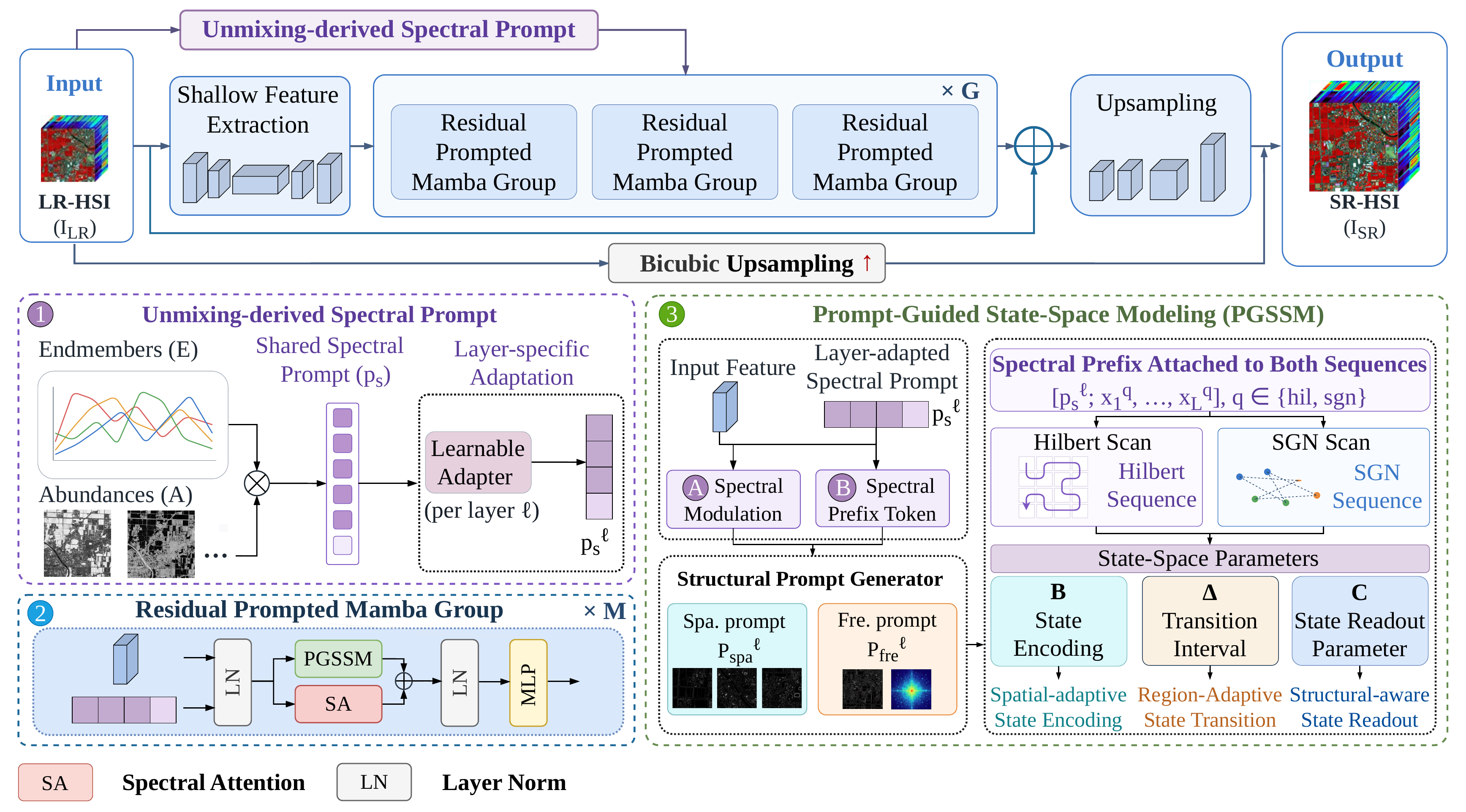} 
\caption{Overall architecture of USP-Mamba.} \label{fig:framework} 
\end{figure*}

The overall architecture of the proposed USP-Mamba is illustrated in Fig.~\ref{fig:framework}. Given a low-resolution hyperspectral image (LR-HSI) $\mathbf{I}_{\mathrm{LR}}\in\mathbb{R}^{H\times W\times C}$, our goal is to reconstruct its high-resolution counterpart $\widehat{\mathbf{I}}_{\mathrm{HR}}\in\mathbb{R}^{rH\times rW\times C}$, where $H$ and $W$ denote the spatial dimensions, $C$ is the number of spectral bands, and $r$ is the upsampling factor. USP-Mamba consists of four major components: an unmixing-derived spectral prompt generator, a shallow feature extraction module, a deep prompt-conditioned state-space backbone, and a high-resolution reconstruction module.

To derive input-specific spectral guidance, the LR-HSI is first processed by a spectral unmixing module: 
\begin{equation} \left( \mathbf{p}_{s}, \widetilde{\mathbf{I}}_{\mathrm{LR}}, \mathbf{E}, \mathbf{A} \right) = \mathcal{U} \left( \mathbf{I}_{\mathrm{LR}} \right), \label{eq:overall_unmixing} 
\end{equation} 
where $\mathbf{E}$ and $\mathbf{A}$ denote the estimated endmember signatures and abundance maps, respectively, and $\widetilde{\mathbf{I}}_{\mathrm{LR}}$ is the corresponding unmixing reconstruction. The resulting spectral prompt $\mathbf{p}_{s}$ summarizes the global composition of the input HSI and provides persistent conditioning for deep feature modeling. Meanwhile, a $3\times3$ convolution extracts the shallow feature: 
\begin{equation} \mathbf{F}_{0} = \mathrm{Conv}_{3\times3} \left( \mathbf{I}_{\mathrm{LR}} \right), 
\label{eq:shallow_feature} 
\end{equation} 
where $\mathbf{F}_{0}\in\mathbb{R}^{H\times W\times D}$ and $D$ denotes the feature dimension.

The deep feature extraction backbone comprises $G$ residual prompt-conditioned state-space groups. The feature propagation through the $g$-th group is formulated as 
\begin{equation} \mathbf{F}_{g} = \mathbf{F}_{g-1} + \mathcal{H}_{g} \left( \mathbf{F}_{g-1}; \mathbf{p}_{s} \right), \qquad g=1,2,\ldots,G, \label{eq:residual_group} 
\end{equation} 
where $\mathcal{H}_{g}(\cdot)$ denotes the $g$-th residual state-space group. Each group contains multiple prompted Mamba blocks followed by a convolutional projection. The group-level residual connection preserves low-frequency information and facilitates stable deep feature learning. For the $m$-th prompted Mamba block in the $g$-th group, the input feature $\mathbf{F}_{g,m-1}$ is updated as 
\begin{equation} 
\begin{split} 
\mathbf{p}_{s}^{g,m} &= {A}_{g,m} \left( \mathbf{p}_{s} \right),\\
\left( \mathbf{P}_{\mathrm{spa}}^{g,m}, \mathbf{P}_{\mathrm{fre}}^{g,m} \right) &= {Q}_{g,m} \left( \mathbf{F}_{g,m-1} \right),\\ 
\mathbf{F}_{g,m} &= {M}_{g,m} \left( \mathbf{F}_{g,m-1}; \mathbf{p}_{s}^{g,m}, \mathbf{P}_{\mathrm{spa}}^{g,m}, \mathbf{P}_{\mathrm{fre}}^{g,m} \right), 
\end{split} 
\label{eq:prompted_block} 
\end{equation} 
where ${A}_{g,m}(\cdot)$ adapts the global spectral prompt to the current feature representation, and ${Q}_{g,m}(\cdot)$ generates the spatial and frequency prompts from the intermediate feature. ${M}_{g,m}(\cdot)$ denotes the proposed prompted Mamba block, in which the spectral prompt provides composition-aware global conditioning, while the structural prompts modulate state-space modeling according to local image characteristics. 

After all residual groups, the deep representation is integrated with the shallow feature through a long residual connection. Finally, the fused feature is projected and spatially enlarged by a PixelShuffle-based reconstruction module.


\subsection{Unmixing-Derived Spectral Prompt}
\label{sec:spectral_prompt}
Hyperspectral pixels are generally composed of mixtures of several latent
materials. We exploit this intrinsic property to derive an input-specific
spectral prompt instead of relying on freely learned prompt parameters.
Given an LR-HSI $\mathbf{I}_{\mathrm{LR}}\in\mathbb{R}^{H\times W\times C}$,
the spectral unmixing branch estimates $K$ endmember signatures $\mathbf{E}\in\mathbb{R}^{K\times C}$ and the corresponding abundance maps $\mathbf{A}\in\mathbb{R}^{H\times W\times K}$. The linear mixing process is formulated as
\begin{equation}
    \widetilde{\mathbf{I}}_{\mathrm{LR}}(i,j,:)
    =
    \sum_{k=1}^{K}
    \mathbf{A}(i,j,k)\mathbf{E}(k,:),
    \label{eq:linear_unmixing}
\end{equation}
subject to
\begin{equation}
    \mathbf{A}(i,j,k)\geq 0,
    \qquad
    \sum_{k=1}^{K}\mathbf{A}(i,j,k)=1.
    \label{eq:abundance_constraint}
\end{equation}

In practice, the abundance constraint is imposed using a channel-wise softmax:
\begin{equation}
    \mathbf{A}
    =
    \mathrm{Softmax}_{K}
    \left(
    \mathrm{Conv}_{A}
    \left(
    \mathbf{I}_{\mathrm{LR}}
    \right)
    \right).
    \label{eq:abundance_estimation}
\end{equation}

To summarize the material composition of the entire image, the abundance maps are spatially aggregated as
\begin{equation}
    \overline{\mathbf{a}}_{k}
    =
    \frac{1}{HW}
    \sum_{i=1}^{H}
    \sum_{j=1}^{W}
    \mathbf{A}(i,j,k),
    \qquad k=1,\ldots,K.
    \label{eq:global_abundance}
\end{equation}
The global spectral prior is then obtained by abundance-weighted aggregation of the endmembers:
\begin{equation}
    \mathbf{v}_{s}
    =
    \sum_{k=1}^{K}
    \overline{\mathbf{a}}_{k}
    \mathbf{E}(k,:),
    \qquad
    \mathbf{p}_{s}
    =
    \mathrm{MLP}_{s}
    \left(
    \mathbf{v}_{s}
    \right),
    \label{eq:spectral_prompt}
\end{equation}
where $\mathbf{p}_{s}\in\mathbb{R}^{D}$ denotes the spectral prompt.
Unlike a conventional learnable prompt shared by all samples,
$\mathbf{p}_{s}$ is explicitly derived from the material composition of
each input.

As shallow and deep layers encode different levels of spectral information,
directly sharing an unchanged prompt across all blocks may constrain its
representation capacity. Therefore, we associate the $l$-th block with a
learnable depth embedding $\mathbf{e}_{l}$ and adapt the prompt as
\begin{equation}
    \mathbf{p}_{s}^{l}
    =
    \mathbf{p}_{s}
    +
    \alpha_{l}
    \mathrm{MLP}_{l}
    \left(
    \mathrm{CAT} \left( \mathbf{p}_{s} ,\mathbf{e}_{l}\right)
    \right),
    \label{eq:layer_prompt}
\end{equation}
where $\mathrm{CAT}(\cdot)$ denotes concatenation and $\alpha_{l}$ is a learnable residual scale. The adapted prompt is prepended to the scanned feature sequence. In this way, the material composition is introduced as a persistent condition for state propagation while remaining responsive to the representation depth.

\subsection{Image-Dependent Structural Prompts}

\label{sec:structural_prompt}

While the spectral prompt establishes composition-aware global conditioning, it does not explicitly account for the heterogeneous structural characteristics across spatial regions. Consequently, regions with distinct geometric and frequency patterns may receive insufficiently differentiated state encoding and transition behaviors~\cite{LC2024, WH2026}. To address this limitation, we introduce spatial and frequency prompts that condition the state-space parameters on image-dependent structural priors.

Given an intermediate feature $\mathbf{F}_{l}\in\mathbb{R}^{H\times W\times D}$, we construct a spatial prompt $\mathbf{P}_{\mathrm{spa}}^{l}$ and a frequency prompt $\mathbf{P}_{\mathrm{fre}}^{l}$ to encode complementary structural characteristics. The spatial prompt is generated from local spatial responses using a $7\times7$ convolution. It preserves position-dependent geometric cues and emphasizes boundaries and fine details without altering the spatial resolution.

To characterize regional frequency variations, we transform the feature into the frequency domain:

\begin{equation}
    \widehat{\mathbf{F}}_{l}
    = \mathrm{FFT}_{2}
    \left(
    \mathbf{F}_{l}
    \right),
    \label{eq:frequency_transform}
\end{equation}
where $\widehat{\mathbf{F}}_{l}$ denotes the complex-valued frequency representation. A learnable complex filter $\mathbf{W}_{f}$ is then applied to recalibrate the frequency responses:

\begin{equation}
    \widehat{\mathbf{F}}_{l}^{\,\prime}
    = \mathbf{W}_{f}
    \odot
    \widehat{\mathbf{F}}_{\ell},
    \label{eq:frequency_filtering}
\end{equation}
where $\odot$ denotes element-wise complex multiplication. The enhanced frequency representation is mapped back to the spatial domain to obtain the frequency prompt:
\begin{equation}
    \mathbf{P}_{\mathrm{fre}}^{l}
    =
    \sigma
    \left(
    \mathrm{Conv}
    \left[
    \mathrm{IFFT}_{2}
    \left(
    \widehat{\mathbf{F}}_{l}^{\,\prime}
    \right)
    \right]
    \right),
    \label{eq:frequency_prompt}
\end{equation}
where $\mathrm{Conv}(\cdot)$ denotes feature projection and $\sigma(\cdot)$ is the sigmoid function. The resulting prompt captures varying frequency responses, enabling the state-space model to distinguish homogeneous regions from high-frequency details.

\begin{table*}[t!]
\centering
\setlength{\tabcolsep}{0.003\hsize}
\begin{tabular}{lccccccccccc}
\toprule[1pt]
\multirow{2}{*}{Method}&\multirow{2}{*}{Scale}&\multicolumn{5}{c}{Chikusei}&\multicolumn{5}{c}{Houston2018}\\
\cmidrule(lr){3-7}\cmidrule(lr){8-12}
&&PSNR$\uparrow$&SSIM$\uparrow$&SAM$\downarrow$&CC$\uparrow$&ERGAS$\downarrow$&PSNR$\uparrow$&SSIM$\uparrow$&SAM$\downarrow$&CC$\uparrow$&ERGAS$\downarrow$\\
\midrule
SSPSR~\cite{SSPSR}&$\times4$&\underline{39.9797}&0.9393&2.4864&\underline{0.9528}&5.1905&45.6017&0.9778&1.9650&0.9850&2.1380\\
RFSR~\cite{RFSR}&$\times4$&39.8950&0.9382&\underline{2.4656}&0.9517&5.2334&45.8677&0.9792&\underline{1.8304}&0.9858&2.0659\\
AS$^{3}$ITransUNet~\cite{AS3Net}&$\times4$&39.9093&0.9377&2.6056&0.9519&\underline{5.1900}&45.8819&0.9792&1.8679&0.9862&2.0731\\
MambaIRv2~\cite{mambairv2}&$\times4$&39.9457&\underline{0.9408}&2.6028&0.9510&5.3115&\underline{46.0946}&\underline{0.9801}&1.8762&\underline{0.9867}&\underline{2.0199}\\
VolFormer~\cite{volformer}&$\times4$&39.8584&0.9371&2.5715&0.9513&5.2263&45.8143&0.9790&1.9121&0.9857&2.0837\\
MambaHSISR~\cite{mambahsisr}&$\times4$&39.4123&0.9321&2.6833&0.9461&5.5534&45.6210&0.9777&1.9185&0.9851&2.1344\\
USP-Mamba &$\times4$ &\textbf{40.2282} &\textbf{0.9433} &\textbf{2.3779} &\textbf{0.9551} &\textbf{5.0462} &\textbf{46.4433} &\textbf{0.9822} &\textbf{1.8043} &\textbf{0.9876} &\textbf{1.9310}\\
\midrule
SSPSR~\cite{SSPSR}&$\times8$&35.1643&0.8299&4.6911&0.8560&9.0504&39.2844&0.9164&4.2673&0.9346&4.4212\\
RFSR~\cite{RFSR}&$\times8$&35.5049&0.8405&\underline{4.2785}&0.8661&\underline{8.6338}&39.4899&0.9211&\underline{3.8403}&0.9379&4.2967\\
AS$^{3}$ITransUNet~\cite{AS3Net}&$\times8$&35.4999&0.8408&4.4746&0.8661&8.6793&\underline{39.8186}&\underline{0.9254}&3.9035&\underline{0.9422}&\underline{4.1466}\\
MambaIRv2~\cite{mambairv2}&$\times8$&35.4818&\underline{0.8456}&4.3894&0.8653&8.7392&39.6079&0.9228&3.8853&0.9403&4.2485\\
VolFormer~\cite{volformer}&$\times8$&\underline{35.5316}&0.8441&4.3289&\underline{0.8674}&8.6549&39.4012&0.9177&4.2017&0.9366&4.3695\\
MambaHSISR~\cite{mambahsisr}&$\times8$&35.1723&0.8328&4.6375&0.8543&9.0137&39.1797&0.9117&4.2231&0.9322&4.4547\\
USP-Mamba&$\times8$&\textbf{35.6353}&\textbf{0.8486}&\textbf{4.1261}&\textbf{0.8709}&\textbf{8.5211}&\textbf{39.9211}&\textbf{0.9255}&\textbf{3.6124}&\textbf{0.9439}&\textbf{4.1293}\\
\bottomrule[1pt]
\end{tabular}
\caption{Quantitative comparison on the Chikusei and Houston2018 datasets at scale factors $\times4$ and $\times8$. The best and second-best results are highlighted in bold and underlined, respectively.}
\label{tab:quantitative_comparison}
\end{table*}

\begin{figure*}[t] 
\centering \includegraphics[width=0.9\textwidth] {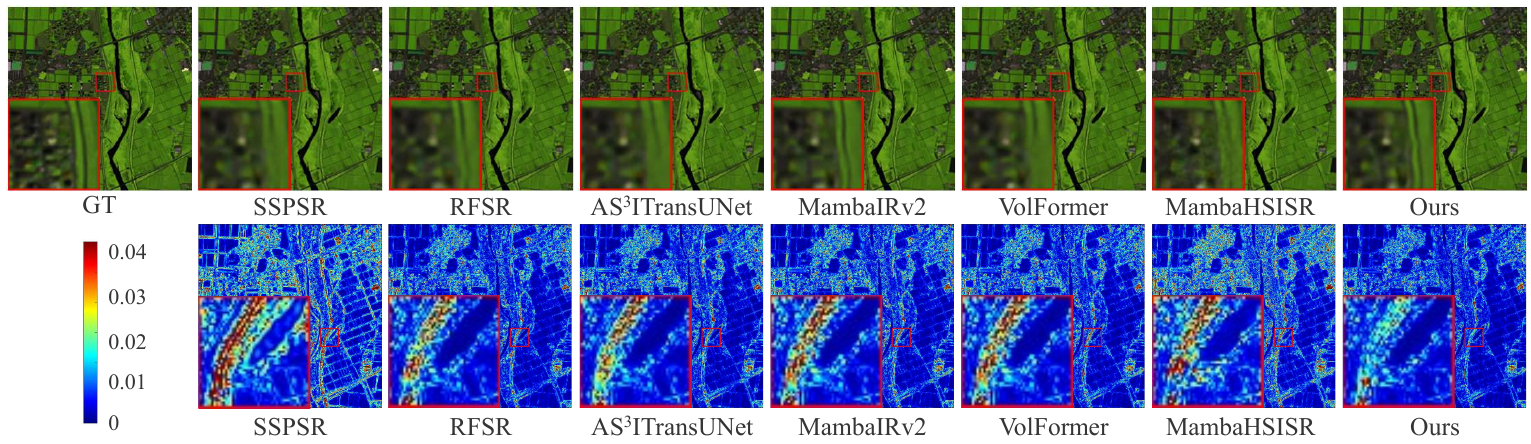} 
\caption{Visual comparison on the Chikusei dataset at scale factor $\times4$. Spectral bands 70, 100, and 36 are displayed as R, G, and B, respectively. The upper row shows reconstructed RGB composites and enlarged regions, while the lower row shows mean error maps across all spectral bands. Bluer regions indicate lower reconstruction errors.} 
\label{fig:chikusei_visual} 
\end{figure*}

\subsection{Prompt-Guided State Space Modeling}

The causal dependencies modeled by Mamba are inherently influenced by the sequence ordering. Conventional row-wise scanning may disrupt spatial continuity and weaken interactions among distant yet semantically related regions. We therefore employ complementary Hilbert and Semantic-Guided Neighboring scans to construct sequences that preserve local geometry and capture non-local semantic dependencies, respectively. 

Let $\mathbf{F}_{\ell}\in\mathbb{R}^{H\times W\times D}$ denote the intermediate feature at the $l$-th block, where $L=HW$ is the number of spatial positions and $D$ is the feature dimension. Let $(u_t,v_t)$ denote the spatial position visited at step $t$ along the Hilbert curve. The corresponding Hilbert sequence is given by 
\begin{equation} 
\mathbf{x}^{\mathrm{hil}}_{t} = \mathbf{F}_{\ell}(u_t,v_t,:), \qquad t=1,\ldots,L. 
\label{eq:hilbert_scan} 
\end{equation}

By preserving the proximity of neighboring pixels in the sequential domain, Hilbert scanning facilitates the propagation of local geometric information.

Following MambaIRV2~\cite{mambairv2}, we adopt Semantic-Guided Neighboring (SGN) scanning to capture non-local semantic dependencies. Given the flattened feature $\mathbf{X}_{l}\in\mathbb{R}^{L\times D}$, the semantic index of the $n$-th token is obtained as 
\begin{equation} 
g_n = \arg\max_k \mathrm{Softmax} \left( \mathrm{Linear} \left( \mathbf{X}_{\ell} \right) \right)_{n,k}. 
\label{eq:semantic_index} 
\end{equation} 

The tokens are then reordered according to their semantic indices: \begin{equation} 
\mathbf{x}^{\mathrm{sgn}}_{t} = \mathbf{X}_{l} \left[ \mathrm{Argsort}(\mathbf{g})_t,: \right], \qquad t=1,\ldots,L, \label{eq:sgn_scan} 
\end{equation} 
where $\mathbf{g}=[g_1,\ldots,g_L]$. This rearrangement places semantically related pixels at nearby sequence positions even when they are spatially distant.

For notational consistency, we use $q\in\{\mathrm{hil},\mathrm{sgn}\}$ to index the Hilbert and SGN scanning strategies, respectively. The layer-adapted spectral prompt is prepended to the sequence generated by each strategy:
\begin{equation} 
\widetilde{\mathbf{x}}^{q} = \left[ \mathbf{p}_{s}^{l}; \mathbf{x}^{q}_{1}, \ldots, \mathbf{x}^{q}_{L} \right], \qquad q\in\{\mathrm{hil},\mathrm{sgn}\},
\label{eq:scan_prompt_prefix} 
\end{equation}
where $\mathbf{p}_{s}^{l}$ integrates a shared composition-aware component with layer-specific spectral prompts, providing shared and layer-specific conditioning for both scanning sequences before state-space modeling.

The input-dependent state-space parameters are predicted from the current feature: \begin{equation} \left( \mathbf{B}_{l}, \mathbf{C}_{l}, \boldsymbol{\Delta}_{l} \right) = \mathrm{Split} \left( \mathrm{Linear} \left( \mathbf{F}_{l} \right) \right), 
\label{eq:ssm_parameters} 
\end{equation} 
where $\mathbf{B}_{l}$ controls state encoding, $\boldsymbol{\Delta}_{l}$ determines the input-dependent transition interval, and $\mathbf{C}_{l}$ performs selective state readout. Before modulation, these parameter maps and the structural prompts are arranged according to the same scanning order. Their branch-aligned forms are denoted by $\mathbf{B}_{l}^{q}$, $\mathbf{C}_{l}^{q}$, $\boldsymbol{\Delta}_{l}^{q}$, $\mathbf{P}_{\mathrm{spa}}^{l,q}$, and $\mathbf{P}_{\mathrm{fre}}^{l,q}$.

The spatial prompt modulates the state encoding parameter: \begin{equation} 
\widehat{\mathbf{B}}_{l}^{q} = \mathbf{B}_{l}^{q} \odot \left[ \mathbf{1} + \alpha_{B} \mathrm{Proj} \left( \mathbf{P}_{\mathrm{spa}}^{l,q} \right) \right],
\label{eq:spatial_prompt_b} 
\end{equation} 
where $\alpha_{B}$ is a learnable scaling factor, and $\mathrm{Proj}(\cdot)$ maps the spatial prompt to the dimensionality of $\mathbf{B}_{l}^{q}$. This modulation enables structure-adaptive state encoding, allowing informative local details to be incorporated more effectively into the latent state.

The frequency prompt adjusts the transition interval: \begin{equation} 
\widehat{\boldsymbol{\Delta}}_{l}^{q} = \mathrm{Softplus} \left[ \boldsymbol{\Delta}_{l}^{q} + \alpha_{\Delta} \mathrm{Proj} \left( \mathbf{P}_{\mathrm{fre}}^{l,q} \right) \right],
\label{eq:frequency_prompt_delta} 
\end{equation} 
where $\alpha_{\Delta}$ controls the modulation strength. This modulation adjusts the state update rate according to local frequency characteristics, producing smoother evolution in homogeneous regions and more responsive updates around high-frequency details.

\begin{figure*}[t] 
\centering 
\includegraphics[width=0.9\textwidth] {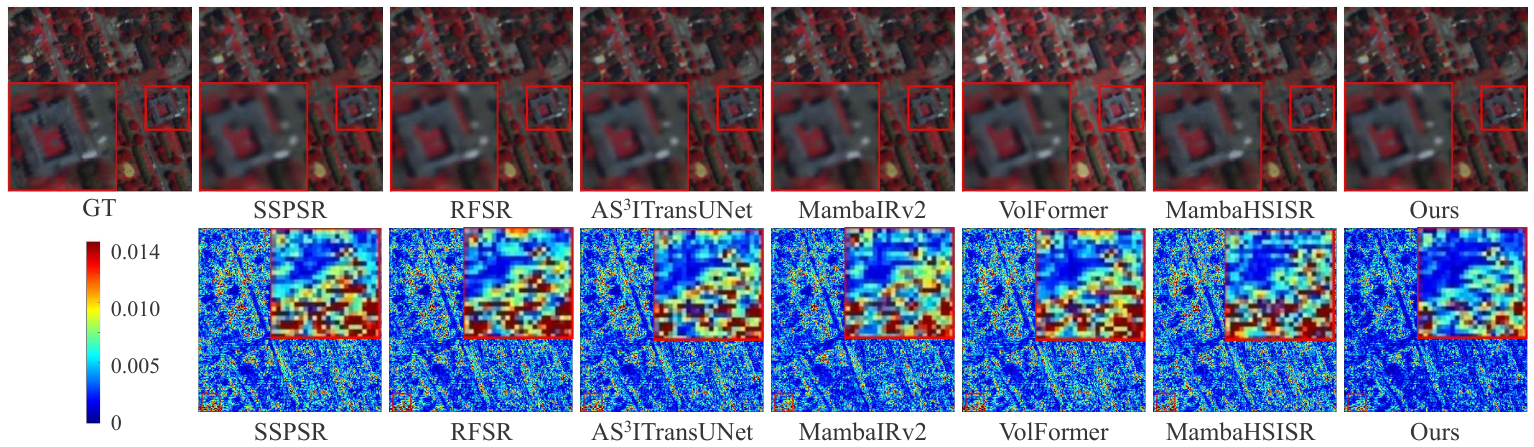} \caption{Visual comparison on the Houston2018 dataset at scale factor $\times4$. Spectral bands 26, 20, and 10 are displayed as R, G, and B, respectively. The upper row shows reconstructed RGB composites and enlarged regions, while the lower row shows mean error maps across all spectral bands. Bluer regions indicate lower reconstruction errors.} 
\label{fig:houston_visual} 
\end{figure*}

The spatial and frequency prompts further condition the state readout: 
\begin{equation} 
\begin{split} 
\widehat{\mathbf{C}}_{l}^{q} = \mathbf{C}_{l}^{q} &+ \alpha_{C}^{\mathrm{spa}} \mathrm{Proj}^{\mathrm{spa}} \left( \mathbf{P}_{\mathrm{spa}}^{l,q} \right)\\ &+ \alpha_{C}^{\mathrm{fre}} \mathrm{Proj}^{\mathrm{fre}} \left( \mathbf{P}_{\mathrm{fre}}^{l,q} \right),
\end{split} 
\label{eq:structural_prompt_c} 
\end{equation}
enabling the latent states to be selectively decoded according to local geometric and frequency characteristics.

The spectral prefix is processed first to establish a composition-conditioned initial state $\mathbf{h}_{0}^{q}$. The subsequent state-space recurrence is written as 
\begin{equation} 
\begin{split} 
\mathbf{h}^{q}_{t} &= \overline{\mathbf{A}}^{q}_{t} \mathbf{h}^{q}_{t-1} + \overline{\mathbf{B}}^{q}_{t} \mathbf{x}^{q}_{t},\\ \mathbf{y}^{q}_{t} &= \widehat{\mathbf{C}}^{q}_{t} \mathbf{h}^{q}_{t} + \mathbf{D} \mathbf{x}^{q}_{t}, 
\end{split} \qquad t=1,\ldots,L, 
\label{eq:prompted_ssm} 
\end{equation} 
where $\overline{\mathbf{A}}^{q}_{t}$ and $\overline{\mathbf{B}}^{q}_{t}$ are discretized using the modulated transition interval $\widehat{\boldsymbol{\Delta}}^{q}_{t}$ and the encoding parameter $\widehat{\mathbf{B}}^{q}_{t}$, respectively. In this manner, the spatial prompt regulates the incorporation of local structural information into latent states, whereas the frequency prompt adjusts state transition rates according to local frequency characteristics.

After reversing the corresponding scanning order, the Hilbert and SGN sequences are restored as spatial feature maps $\mathbf{Y}_{l}^{\mathrm{hil}}$ and $\mathbf{Y}_{l}^{\mathrm{sgn}}$, respectively. The two representations are then aggregated through channel concatenation and linear projection.

\subsection{Loss Function} \label{sec:loss} Following previous HSISR methods~\cite{CW2023, HW2024}, we employ the $\ell_1$ reconstruction loss, spectral angle loss, and gradient loss to supervise spatial--spectral reconstruction. Given the reconstructed HR-HSI $\widehat{\mathbf{I}}$ and its ground truth $\mathbf{I}_{\mathrm{gt}}$, the overall objective is defined as 
\begin{equation} \mathcal{L} = \mathcal{L}_{1} + \lambda_{\mathrm{sam}}\mathcal{L}_{\mathrm{sam}} + \lambda_{\mathrm{grad}}\mathcal{L}_{\mathrm{grad}} + \lambda_{\mathrm{unm}}\mathcal{L}_{\mathrm{unm}}, 
\label{eq:total_loss} 
\end{equation} 
where $\mathcal{L}_{1}$, $\mathcal{L}_{\mathrm{sam}}$, and $\mathcal{L}_{\mathrm{grad}}$ denote the reconstruction, spectral angle, and gradient losses, respectively. To constrain the unmixing branch, we further minimize the difference between the spectrally normalized input LR-HSI and its unmixing reconstruction.


\section{Experiments}
\subsection{Datasets} \label{sec:datasets} We conduct experiments on two remote-sensing hyperspectral datasets, including Chikusei~\cite{chikusei} and Houston2018. The Chikusei dataset was acquired over agricultural and urban areas in Japan. It contains 128 spectral bands and has a spatial size of $2517\times2335$. Following HSRMamba~\cite{hsrmamba}, four non-overlapping cubes of size $512\times512\times128$ are cropped from the upper region for testing, while the remaining area is used for training and validation. Houston 2018 was collected over the University of Houston and the surrounding urban areas. It contains 48 spectral bands with a spatial size of $4172\times1202$. Eight non-overlapping cubes of size $256\times256\times48$ are selected from the upper region for testing, and the remaining region is used for training and validation. For both datasets, LR-HSIs are generated from the corresponding HR-HSIs using bicubic downsampling at scale factors $\times4$ and $\times8$. During training, the LR patch size is set to $32\times32$, corresponding to HR patches of $128\times128$ and $256\times256$, respectively.

\subsection{Implementation details}
The feature dimension and number of endmembers are set to 64 and 16, respectively, with a batch size of 16. The numbers of prompted Mamba blocks in the four stages are configured as [2, 2, 2, 2]. The loss weights $\lambda_{\mathrm{sam}}$, $\lambda_{\mathrm{grad}}$, and $\lambda_{\mathrm{unm}}$ are empirically set to $0.1$, $0.1$, and $0.005$, respectively. The model is optimized using Adam for 300 epochs with an initial learning rate of $5\times10^{-5}$. A cosine annealing schedule is adopted to gradually reduce the learning rate to $2.5\times10^{-5}$. All experiments are implemented in PyTorch and conducted on NVIDIA GeForce RTX 4090 GPU. We compare USP-Mamba with seven representative SHSR methods, including the CNN-based SSPSR~\cite{SSPSR} and RFSR~\cite{RFSR}; the Transformer-based AS$^{3}$ITransUNet~\cite{AS3Net}, and VolFormer~\cite{volformer}; and the Mamba-based MambaIRv2~\cite{mambairv2} and MambaHSISR~\cite{mambahsisr}. Reconstruction quality is evaluated using five commonly adopted spatial and spectral metrics: peak signal-to-noise ratio (PSNR), structure similarity (SSIM), spectral angle mapper (SAM), cross-correlation (CC), and erreur relative global adimensionnellede synthese (ERGAS). Higher PSNR, SSIM, and CC values indicate better performance, while lower SAM and ERGAS values are preferred.

\subsection{Results on the Chikusei Dataset} \label{sec:chikusei_results} The quantitative results on Chikusei and Houston2018 are jointly reported in Table~\ref{tab:quantitative_comparison}.

On Chikusei, USP-Mamba achieves a PSNR of 40.2282 dB at $\times4$, outperforming the second-best method by 0.2485 dB. At the more challenging $\times8$ scale, it also obtains the lowest SAM of 4.1261. Consistent improvements are observed across the remaining metrics, including SSIM, CC, and ERGAS, indicating that the proposed method maintains a favorable balance between spatial reconstruction and spectral preservation. This performance can be attributed to the unmixing-derived spectral prompt, which introduces composition-aware information into state propagation, together with the spatial and frequency prompts that adapt state modeling to local geometry and frequency variations. The complementary Hilbert and SGN scanning paths enhance local continuity and non-local dependency modeling.

Figure~\ref{fig:chikusei_visual} presents the visual comparison at $\times4$. The compared methods exhibit varying degrees of boundary smoothing and residual artifacts in regions containing narrow field boundaries and fine textures. In contrast, USP-Mamba reconstructs sharper geometric details and produces results visually closer to the ground truth. The corresponding mean error maps further demonstrate that USP-Mamba produces the lowest overall reconstruction error among all compared methods.

\begin{figure}[t] 
\centering 
\includegraphics[width=0.95\columnwidth] {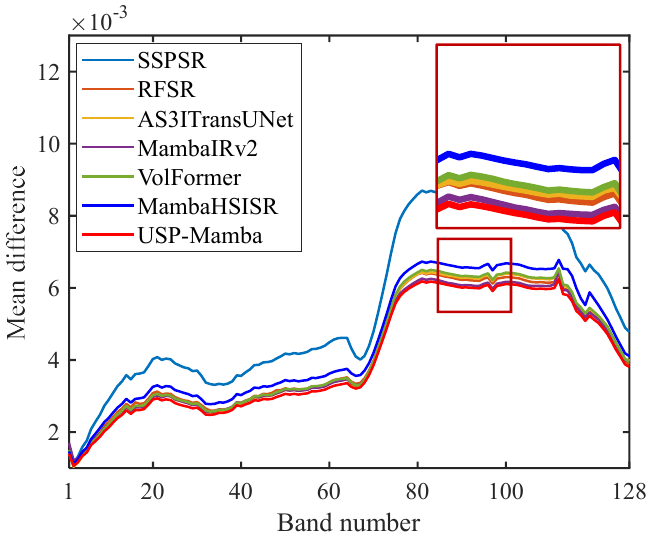} \caption{Mean spectral difference curves of different methods on the Chikusei dataset at scale factor $\times4$.} 
\label{fig:spectral_curve} 
\end{figure}

To further evaluate spectral reconstruction, Fig.~\ref{fig:spectral_curve} compares the mean spectral difference curves of different methods on Chikusei at $\times4$. USP-Mamba maintains a lower spectral difference over most bands, particularly in regions exhibiting larger spectral variations. This observation demonstrates that the unmixing-derived spectral prompt effectively conditions state propagation with composition-aware information and reduces spectral distortion during reconstruction. 

\subsection{Results on the Houston2018 Dataset} 
\label{sec:houston_results} 
The quantitative results on Houston2018 are also provided in Table~\ref{tab:quantitative_comparison}. USP-Mamba achieves a PSNR of 46.4433 dB at $\times4$ and reduces SAM to 3.6124 at $\times8$. It also provides the best overall results across the two scale factors and the remaining evaluation metrics. The improvements suggest that the proposed prompting mechanism remains effective for urban scenes containing diverse materials and dense spatial boundaries. In particular, composition-aware spectral conditioning helps reduce spectral distortion, while structure-adaptive spatial and frequency modulation facilitates the reconstruction of local details under different spatial degradation levels. The combination of Hilbert and SGN scanning provides complementary local and non-local feature propagation.

\subsection{Ablation Study} \label{sec:ablation} We conduct an ablation study on Chikusei at $\times4$ scale factor. All variants are trained using the same configuration. 

\noindent\textbf{Effectiveness of different prompts.} We evaluate the contributions of the unmixing-derived spectral prompt (USP), spatial prompt (SpaP), and frequency prompt (FreP). As shown in Table~\ref{tab:prompt_ablation}, each prompt individually improves PSNR over the baseline, suggesting a stronger ability to preserve spatial structures. Moreover, jointly using SpaP and FreP yields further gains in both PSNR and SSIM. The full model achieves the best overall performance, confirming the complementarity between composition-aware spectral conditioning and structure-adaptive spatial--frequency modulation.

\begin{table}[t]
\centering 
\setlength{\tabcolsep}{4.0pt} 
\begin{tabular}{c c c c c c c} 
\toprule 
Variant & USP & SpaP & FreP & PSNR$\uparrow$ &SSIM$\uparrow$ & SAM$\downarrow$ \\ \midrule 
(a) & & & & 39.8833 & 0.9327 & 2.4880 \\ 
(b) & $\checkmark$ & & & 40.0738 & 0.9329 & 2.4776\\ 
(c) & & $\checkmark$ & & 40.0992 & 0.9386 & 2.5065 \\ 
(d) & & & $\checkmark$ & 40.0764 & 0.9328 & 2.4885 \\ 
(e) & & $\checkmark$ & $\checkmark$ & 40.1338 & 0.9426 & 2.4796 \\ 
(f) & $\checkmark$ & $\checkmark$ & $\checkmark$ & \textbf{40.2282} & \textbf{0.9433}  & \textbf{2.3779}\\ 
\bottomrule
\end{tabular} 
\caption{Ablation study of different prompt components on Chikusei under the $\times4$ setting.}
\label{tab:prompt_ablation} 
\end{table}

\noindent\textbf{Effectiveness of complementary scanning.} We further investigate different scanning combinations while retaining all prompt components and two scanning paths. As shown in Table~\ref{tab:scan_ablation}, replacing conventional raster scanning with Hilbert scanning improves PSNR from 39.9254 dB to 40.0285 dB, indicating the benefit of preserving spatial continuity. The combination of Hilbert and SGN achieves the best performance, demonstrating the effectiveness of these components.
\begin{table}[t] 
\centering 
\begin{tabular}{c c c c} 
\toprule 
Variant & First path & Second path & PSNR (dB) \\ 
\midrule 
(a) & Raster & Reverse & 39.9254 \\ 
(b) & Hilbert & Reverse & 40.0285 \\ 
(c) & Hilbert & SGN & \textbf{40.2282} \\ 
\bottomrule
\end{tabular} 
\caption{Ablation study of scanning strategies on Chikusei under the $\times4$ setting.} 
\label{tab:scan_ablation} 
\end{table}

\section{Conclusion}
In this paper, we proposed USP-Mamba for hyperspectral image super-resolution. The method introduces an unmixing-derived spectral prompt to condition state propagation with material-composition priors, while spatial and frequency prompts adapt state encoding, state transition, and readout to local structural variations. In addition, Hilbert and SGN scanning are combined to model complementary local continuity and non-local dependencies. Extensive experiments on the Chikusei and Houston2018 datasets under $\times4$ and $\times8$ settings demonstrate that USP-Mamba achieves consistently favorable spatial--spectral reconstruction performance compared with representative methods.




\bibliography{aaai2027}

@article{HW2024,
  author={Hu, Qian and Wang, Xinya and Jiang, Junjun and Zhang, Xiao-Ping and Ma, Jiayi},
  journal={IEEE Transactions on Image Processing}, 
  title={Exploring the Spectral Prior for Hyperspectral Image Super-Resolution}, 
  year={2024},
  volume={33},
  pages={5260-5272}
}

@article{CST,
  author       = {Shi Chen and
                  Lefei Zhang and
                  Liangpei Zhang},
  title        = {Cross-scope spatial-spectral information aggregation for hyperspectral image super-resolution},
  journal      = {IEEE Transactions on Image Processing},
  volume       = {33},
  pages        = {5878-5891},
  year         = {2024}
}

@article{WW2025,
  title={Hierarchical context measurement network for single hyperspectral image super-resolution},
  author={Wang, Heng and Wang, Cong and Yuan, Yuan},
  journal={IEEE Transactions on Multimedia},
  volume={27},
  pages={2623--2637},
  year={2025},
}

@article{SSPSR,
  author    = {Junjun Jiang and
               He Sun and
               Xianming Liu and
               Jiayi Ma},
  title     = {Learning Spatial-Spectral Prior for Super-Resolution of Hyperspectral Imagery},
  journal   = {{IEEE} Trans. Computational Imaging},
  volume    = {6},
  pages     = {1082--1096},
  year      = {2020}
}

@article{RFSR,
  author    = {Xinya Wang and
               Jiayi Ma and
               Junjun Jiang},
  title     = {Hyperspectral Image Super-Resolution via Recurrent Feedback Embedding and Spatial-Spectral Consistency Regularization},
  journal   = {IEEE Transactions on Geoscience and Remote Sensing},
  volume    = {60},
  pages     = {1--13},
  year      = {2022}
}

@article{AS3Net,
  author       = {Qin Xu and
                  Shiji Liu and
                  Jiahui Wang and
                  Bo Jiang and
                  Jin Tang},
  title        = {AS\({}^{3}\)ITransUNet: Spatial-Spectral Interactive Transformer U-Net With Alternating Sampling for Hyperspectral Image Super-Resolution},
  journal      = {IEEE Transactions on Geoscience and Remote Sensing},
  volume       = {61},
  pages        = {1--13},
  year         = {2023},
}

@inproceedings{mambair,
  author       = {Hang Guo and
                  Jinmin Li and
                  Tao Dai and
                  Zhihao Ouyang and
                  Xudong Ren and
                  Shu{-}Tao Xia},
  title        = {MambaIR: {A} Simple Baseline for Image Restoration with State-Space Model},
  booktitle    = {Proceedings of the European Conference on Computer Vision},
  volume       = {15076},
  pages        = {222--241},
  year         = {2024},
}

@inproceedings{mambairv2,
  title={Mambairv2: Attentive state space restoration},
  author={Guo, Hang and Guo, Yong and Zha, Yaohua and Zhang, Yulun and Li, Wenbo and Dai, Tao and Xia, Shu-Tao and Li, Yawei},
  booktitle={Proceedings of the {IEEE/CVF} Computer Vision and Pattern Recognition Conference},
  pages={28124--28133},
  year={2025}
}

@inproceedings{volformer,
  title={VolFormer: Explore more comprehensive cube interaction for hyperspectral image restoration and beyond},
  author={Yu, Dabing and Gao, Zheng},
  booktitle={Proceedings of the {IEEE/CVF} Computer Vision and Pattern Recognition Conference},
  pages={28091--28101},
  year={2025}
}

@article{mambahsisr,
  author={Xu, Yinghao and Wang, Hao and Zhou, Fei and Luo, Chunbo and Sun, Xin and Rahardja, Susanto and Ren, Peng},
  journal={IEEE Transactions on Geoscience and Remote Sensing}, 
  title={MambaHSISR: Mamba Hyperspectral Image Super-Resolution}, 
  year={2025},
  volume={63},
  pages={1-16},
}

@inproceedings{CL2022,
  author       = {Yuanhao Cai and
                  Jing Lin and
                  Xiaowan Hu and
                  Haoqian Wang and
                  Xin Yuan and
                  Yulun Zhang and
                  Radu Timofte and
                  Luc Van Gool},
  title        = {Mask-guided Spectral-wise Transformer for Efficient Hyperspectral Image Reconstruction},
  booktitle    = {Proceedings of the {IEEE/CVF} Conference on Computer Vision and Pattern Recognition},
  pages        = {17481--17490},
  year         = {2022},
}

@inproceedings{SSM,
  author       = {Albert Gu and
                  Karan Goel and
                  Christopher R{\'{e}}},
  title        = {Efficiently Modeling Long Sequences with Structured State Spaces},
  booktitle    = {Proceedings of the International Conference on Learning Representations},
  year         = {2022},
}

@article{mamba,
  title={Mamba: Linear-time sequence modeling with selective state spaces},
  author={Gu, Albert and Dao, Tri},
  journal={arXiv preprint arXiv: 2312.00752},
  year={2023}
}

@inproceedings{hsrmamba,
  title={HSRMamba: Contextual Spatial-Spectral State Space Model for Single Hyperspectral Image Super-Resolution},
  author={Chen, Shi and Zhang, Lefei and Zhang, Liangpei},
  booktitle={Proceedings of the 34th International Joint Conference on Artificial Intelligence},
  year    = {2025},
  pages={810–818}, 
}

@article{sqformer,
  title={SQformer: Spectral-query transformer for hyperspectral image arbitrary-scale super-resolution},
  author={Jiang, Shuguo and Li, Nanying and Xu, Meng and Zhang, Shuyu and Jia, Sen},
  journal={IEEE Transactions on Geoscience and Remote Sensing},
  volume={62},
  pages={1--15},
  year={2024},
}

@inproceedings{essaformer,
  author       = {Mingjin Zhang and
                  Chi Zhang and
                  Qiming Zhang and
                  Jie Guo and
                  Xinbo Gao and
                  Jing Zhang},
  title        = {ESSAformer: Efficient Transformer for Hyperspectral Image Super-resolution},
  booktitle    = {Proceedings of the {IEEE/CVF} International Conference on Computer Vision},
  year         = {2023}
}

@article{msdformer,
  author       = {Shi Chen and
                  Lefei Zhang and
                  Liangpei Zhang},
  title        = {MSDformer: Multiscale Deformable Transformer for Hyperspectral Image Super-Resolution},
  journal      = {IEEE Transactions on Geoscience and Remote Sensing},
  volume       = {61},
  pages        = {1--14},
  year         = {2023},
}

@inproceedings{su2025eigensr,
  title={{EigenSR}: Eigenimage-Bridged Pre-Trained RGB Learners for Single Hyperspectral Image Super-Resolution},
  author={Su, Xi and Shen, Xiangfei and Wan, Mingyang and Nie, Jing and Chen, Lihui and Liu, Haijun and Zhou, Xichuan},
  booktitle={Proceedings of the 39th AAAI Conference on Artificial Intelligence},
  pages={7033--7041},
  year={2025},
}

@inproceedings{LQ2025,
  title={Breaking the Spatial-Temporal Consistency Constraint: Towards Reference-Based Hyperspectral Image Super-Resolution},
  author={Liu, Xuyao and Qu, Jiahui and Dong, Wenqian},
  booktitle={Proceedings of the 33rd ACM International Conference on Multimedia},
  pages={2094--2103},
  year={2025}
}

@inproceedings{ZL2026,
  title={M3SR: Multi-Scale Multi-Perceptual Mamba for Efficient Spectral Reconstruction},
  author={Zhang, Yuze and Li, Lingjie and Lin, Qiuzhen and Ming, Zhong and Yu, Fei and Leung, Victor CM},
  booktitle={Proceedings of the 40th AAAI Conference on Artificial Intelligence},
  pages={12979--12987},
  year={2026}
}

@inproceedings{JW2026,
  title={MFmamba: A multi-function network for panchromatic image resolution restoration based on state-space model},
  author={Jiang, Qian and Wang, Qianqian and Jin, Xin and Wo{\'z}niak, Micha{\l} and Yao, Shaowen and Zhou, Wei},
  booktitle={Proceedings of the 40th AAAI Conference on Artificial Intelligence},
  pages={5406--5414},
  year={2026}
}

@article{fmsr,
  title={Frequency-Assisted Mamba for Remote Sensing Image Super-Resolution},
  author={Xiao, Yi and Yuan, Qiangqiang and Jiang, Kui and Chen, Yuzeng and Zhang, Qiang and Lin, Chia-Wen},
  journal={IEEE Transactions on Multimedia},
  volume= {27},
  pages= {1783--1796},
  year={2024}
}

@article{LC2024,
  title={Fourier-enhanced implicit neural fusion network for multispectral and hyperspectral image fusion},
  author={Liang, Yu-Jie and Cao, Zihan and Deng, Shangqi and Dou, Hong-Xia and Deng, Liang-Jian},
  journal={Advances in neural information processing systems},
  volume={37},
  pages={63441--63465},
  year={2024}
}

@article{chikusei,
  title={Airborne hyperspectral data over Chikusei},
  author={Yokoya, Naoto and Iwasaki, Akira},
  journal={Space Appl. Lab., Univ. Tokyo, Tokyo, Japan, Tech. Rep. SAL-2016-05-27},
  year={2016}
}

@article{CW2023,
  title={A review of hyperspectral image super-resolution based on deep learning},
  author={Chen, Chi and Wang, Yongcheng and Zhang, Ning and Zhang, Yuxi and Zhao, Zhikang},
  journal={Remote Sensing},
  volume={15},
  number={11},
  pages={2853},
  year={2023},
}

@inproceedings{WH2026,
  title={{GEWDiff}: Geometric Enhanced Wavelet-based Diffusion Model for Hyperspectral Image Super-resolution},
  author={Wang, Sirui and He, Jiang and Blasco Andreo, Nat{\`a}lia and Zhu, Xiao Xiang},
  booktitle={Proceedings of the 40th AAAI Conference on Artificial Intelligence},
  pages={10109--10117},
  year={2026}
}

@inproceedings{ZZ2026,
  title={Enhancing Unregistered Hyperspectral Image Super-Resolution via Unmixing-based Abundance Fusion Learning},
  author={Zhang, Yingkai and Zhang, Tao and Nie, Jing and Fu, Ying},
  booktitle={Proceedings of the IEEE/CVF Conference on Computer Vision and Pattern Recognition},
  pages={41573--41583},
  year={2026}
}

@article{XF2026,
  title={Uncertainty-Driven Generative Prior Learning for Sparse Model-Guided Hyperspectral Image Fusion},
  author={Xu, Junwei and Feng, Teng and Fang, Zhenxuan and Wu, Fangfang and Dong, Le and Huang, Tao and Yang, Zhou and Dong, Weisheng and Li, Xin},
  journal={IEEE Transactions on Image Processing},
  year={2026},
}

@inproceedings{XZ2025,
  title={Hipandas: Hyperspectral image joint denoising and super-resolution by image fusion with the panchromatic image},
  author={Xu, Shuang and Zhao, Zixiang and Bai, Haowen and Yu, Chang and Peng, Jiangjun and Cao, Xiangyong and Meng, Deyu},
  booktitle={Proceedings of the IEEE/CVF International Conference on Computer Vision},
  pages={12002--12011},
  year={2025}
}

@inproceedings{PR2024,
  title={Hir-diff: Unsupervised hyperspectral image restoration via improved diffusion models},
  author={Pang, Li and Rui, Xiangyu and Cui, Long and Wang, Hongzhong and Meng, Deyu and Cao, Xiangyong},
  booktitle={Proceedings of the IEEE/CVF conference on computer vision and pattern recognition},
  pages={3005--3014},
  year={2024}
}

@inproceedings{MJ2025,
  author       = {Mengting Ma and
                  Yizhen Jiang and
                  Mengjiao Zhao and
                  Jiaxin Li and
                  Wei Zhang},
  title        = {HetSSNet: Spatial-Spectral Heterogeneous Graph Learning Network for
                  Panchromatic and Multispectral Images Fusion},
  booktitle    = {Proceedings of the 42rd International Conference on Machine Learning},
  year         = {2025},

}

@inproceedings{XF2026b,
  title={TPTransformer: Tensor-Tensor Product Transformer for Hyperspectral Image Super-Resolution},
  author={Xu, Honghui and Fang, Chuangjie and Meng, Yiqun and Jiang, Jiawei and Chan, Sixian and Zhang, Shiqing and Zheng, Jianwei},
  booktitle={Proceedings of the IEEE/CVF Conference on Computer Vision and Pattern Recognition},
  pages={1670--1679},
  year={2026}
}

@inproceedings{XZ2026,
  title={TRT: harnessing tensor ring transformer for hyperspectral image super-resolution},
  author={Xu, Honghui and Zhu, Junwei and Gu, Yubin and Quan, Yueqian and Fang, Chuangjie and Qiu, Hong and Zheng, Jianwei},
  booktitle={Proceedings of the 40th AAAI Conference on Artificial Intelligence},
  pages={11232--11240},
  year={2026}
}


\end{document}